\documentclass[runningheads]{article}
\usepackage[T1]{fontenc} % fonts
\usepackage{graphicx} % images

\usepackage{glossaries} % acronyms definition
\usepackage{color} % used for coloring text
\usepackage{xspace} % space after scapes
\usepackage{amsmath}  % Provides \mathbb for math symbols
\usepackage{amssymb}  % Ensures proper font rendering for \mathbb
\usepackage{amsfonts} % Real set R
\usepackage{bm} % bold font for vectors

\usepackage{subcaption} % subfigures

\usepackage{booktabs}
\newacronym{ai}{AI}{Artificial Inteligence}
\newacronym{aocs}{AOCS}{Attitude and Orbit Control Subsystem}
\newacronym{auc}{AUC}{Area Under the Curve}

\newacronym{cam}{CAM}{Coverage Analysis Method}
\newacronym{cnn}{CNN}{Convolutional Neural Network}
\newacronym{cve}{CVE}{CoreVector Extractor}

\newacronym{dnn}{DNN}{Deep Neural Network}
\newacronym{dum}{DUM}{Deterministic Uncertainty Measurement}

\newacronym{eu}{EU}{European Union}
\newacronym{ead}{EAD}{Explainable Anomaly Detection}

\newacronym{fdir}{FDIR}{Fault Detection, Isolation and Recovery}
\newacronym{fm}{FM}{Feature Mixing}

\newacronym{gmm}{GMM}{Gaussian Mixture Model}
\newacronym{gp}{GP}{Gaussian Process}

\newacronym{hlf}{HLF}{High-Level Feature}

\newacronym{id}{ID}{In-Distribution}

\newacronym{km}{KM}{K-Means}

\newacronym{llf}{LLF}{Low-Level Feature}
\newacronym{lfe}{LFE}{Low-level Feature Extractor}
\newacronym{lrp}{LRP}{Layer-wise Relevance Propagation}

\newacronym{ml}{ML}{Machine Learning}
\newacronym{mse}{MSE}{Mean Squared Error}

\newacronym{rw}{RW}{Reaction Wheel}

\newacronym{svd}{SVD}{Singular Value Decomposition}
\newacronym{snr}{SNR}{Signal-to-Noise Ratio}

\newacronym{xai}{XAI}{eXplainable Artificial Intelligence}

\newacronym{ourframework}{\color{red}Banana\color{black}}{\color{red}We Need a Name\color{black}\xspace}

\newcommand{\codeFont}[1]{\texttt{#1}\xspace}

\newcommand{\realSet}{\mathbb{R}}

\newcommand{\mat}[1]{\ensuremath{\bm{{#1}}}}
\newcommand{\vecc}[1]{\ensuremath{\bm{{#1}}}}

\newcommand{\expect}{\mathbb{E}}

\newcommand{\identity}{\mat{I}}

\newcommand{\img}{\mat{X}}
\newcommand{\numLabels}{\ensuremath{L}}

\newcommand{\weights}{\mat{W}}
\newcommand{\biases}{\mat{b}}
\newcommand{\act}{\vecc{x}}
\newcommand{\actIn}{\act}

\newcommand{\actOut}{\vecc{y}}

\newcommand{\telemetry}{\mat{w}}
\newcommand{\noise}{\mat{\nu}}

\newcommand{\SVDU}{\mat{P}}
\newcommand{\SVDV}{\mat{Q}}
\newcommand{\SVDS}{\mat{\Sigma}}

\newcommand{\means}{\vec{\mu}}
\newcommand{\numClusters}{C}
\newcommand{\covariance}{\mat{K}}
\newcommand{\clusterProb}{\phi}

\newcommand{\corevecToCluster}{\ensuremath{c}}
\newcommand{\corevecToTag}{\ensuremath{g}}
\newcommand{\corevecToClass}{\ell}
\newcommand{\numTags}{\ensuremath{T}}

\newcommand{\affineTransform}{\mat{A}}

\newcommand{\corevector}{\codeFont{corevector}} % textual
\newcommand{\corevectors}{\codeFont{corevectors}} % textual

\newcommand{\coreVec}{\vecc{v}}
\newcommand{\coreVecSize}{\kappa}
\newcommand{\membership}{\vecc{d}}
\newcommand{\membershipElement}{d}
\newcommand{\empiricalPosterior}{\mat{U}}

\newcommand{\empiricalPosteriorElement}{U}
\newcommand{\empiricalPosteriorRowElement}{u}

\newcommand{\peephole}{\codeFont{peephole}} % textual
\newcommand{\peepholes}{\codeFont{peepholes}} % textual
\newcommand{\ph}{\vec{p}}

\begin{document}
\title{On-board Telemetry Monitoring in Autonomous Satellites: Challenges and Opportunities}
%
% \title{On-board monitoring in Autonomous Satellites}
% \author{
%     Lorenzo Capelli \and %\inst{1} \and
%     Leandro de Souza Rosa \and %\inst{1} \and
%     Maurizio De Tommasi \and %\inst{1} \and\\
%     Livia Manovi \and %\inst{1} \and
%     Andriy Enttsel \and %\inst{1} \and  
%     Ilaria Pinci \and %\inst{2} \and
%     Carlo Ciancarelli \and %\inst{2} \and
%     Eleonora Mariotti \and %\inst{3} \and
%     Gianluca Furano \and %\inst{3} \and
%     Mauro Mangia \and %\inst{1} \and
%     Riccardo Rovatti %\inst{1}
% }

\author{
Lorenzo Capelli$^{1}$ \and
Leandro de Souza Rosa$^{1}$ \and
Maurizio De Tommasi$^{1}$ \and
Livia Manovi$^{1}$ \and
Andriy Enttsel$^{1}$ \and
Mauro Mangia$^{1}$ \and
Riccardo Rovatti$^{1}$ \and
Ilaria Pinci$^{2}$ \and
Carlo Ciancarelli$^{2}$ \and
Eleonora Mariotti$^{3}$ \and
Gianluca Furano$^{3}$
\\[6pt]
$^{1}$ University of Bologna, Italy \\[4pt]
\{l.capelli, leandro.desouzarosa, maurizio.detommasi,\\ livia.manovi, andriy.enttsel, mauro.mangia, riccardo.rovatti\}@unibo.it \\[4pt]
$^{2}$ Thales Alenia Space Italia, Rome, Italy \\
\{ilaria.pinci, carlo.ciancarelli\}@thalesaleniaspace.com \\[4pt]
$^{3}$ European Space Agency (ESA-ESTEC), Netherlands \\
\{eleonora.mariotti,gianluca.furano\}@esa.int
}

%
% \authorrunning{L. Capelli et al.}
% \institute{DEI-Arces, University of Bologna, Italy \email{\{l.capelli,~leandro.desouzarosa,~maurizio.detommasi,\\~livia.manovi,~andriy.enttsel,~mauro.mangia,~riccardo.rovatti\}@unibo.it} \and
% TAS-I, Thales Alenia Space Italia S.p.A., Rome, Italy
% \email{\{ilaria.pinci,carlo.ciancarelli\}@thalesaleniaspace.com} \and
% ESA-ESTEC, European Space Agency, Noordwijk, The Netherlands
% \email{\{eleonora.mariotti,gianluca.furano\}@esa.int}}
%
\date{}
\maketitle              % typeset the header of the contribution
\begin{abstract}
	The increasing autonomy of spacecraft demands fault-detection systems that are both reliable and explainable.
    This work addresses \acrlong{xai} for onboard \acrlong{fdir} within the \acrlong{aocs} by introducing a framework that enhances interpretability in neural anomaly detectors.
    We propose a method to derive low-dimensional, semantically annotated encodings from intermediate neural activations, called peepholes.
    Applied to a convolutional autoencoder, the framework produces interpretable indicators that enable the identification and localization of anomalies in reaction-wheel telemetry.
    Peepholes analysis further reveals bias detection and supports fault localization.
    The proposed framework enables the semantic characterization of detected anomalies while requiring only a marginal increase in computational resources, thus supporting its feasibility for on-board deployment.
    %Results highlight that embedding explainability directly within the on-board detection loop can strengthen transparency, verification, and trust in autonomous satellite operations.
	
	% \keywords{Explainable AI \and Telemetry Monitoring \and On-board Fault Detection.}
\end{abstract}

\section{Introduction}
\label{sec: intro}
% The entire space segment is evolving rapidly with an expected impressive growth in terms of mission and market \cite{ESA2024,NASA2024,PlanetPelican2_2024}.
% The ongoing evolution of space missions is redefining the traditional boundaries between ground and onboard operations. Satellites are progressively shifting from passive system relays to active, intelligent systems capable of interpreting telemetry, reacting to anomalies, and making time-critical decisions directly in orbit \cite{furano2020towards}. This paradigm shift, the massive introduction of on-board intelligence, promises to enhance autonomy, reduce latency, and improve resilience against unforeseen conditions \cite{Giuffrida2022,Labreche2022,Mateo-Garcia2023}. However, it also introduces fundamental challenges concerning reliability, certification, and trust. When part of the decision-making process moves from ground control to an autonomous on-board entity, the ability to understand and explain that entity’s behavior becomes as important as its raw performance \cite{nannini2024operationalizing,leveson2011engineering}.

The rapidly evolving space segment is expected to grow in mission scope and market size, redefining the boundaries between ground and onboard operations \cite{ESA2024,NASA2024,PlanetPelican2_2024}. Satellites are progressively shifting from passive to active systems, capable of interpreting telemetry, reacting to anomalies, and making time-critical decisions directly in orbit \cite{furano2020towards}. 
This paradigm shift, marked by the massive introduction of on-board intelligence, promises to enhance autonomy, reduce latency, and improve resilience against unforeseen conditions \cite{Giuffrida2022,Labreche2022,Mateo-Garcia2023}. However, it introduces fundamental challenges concerning reliability, certification, and trust, as understanding and explaining the on-board autonomous decision-making process is as important as its raw performance \cite{nannini2024operationalizing,leveson2011engineering}.

Among the various spacecraft subsystems, we consider a \gls{fdir} block as a strong motivating example as it represents a particularly relevant case to study autonomy. In conventional architectures, \gls{fdir} logic relies on predefined thresholds, confirmation times, and operator supervision \cite{olive2012fdi}. While effective for known failure modes, this deterministic approach cannot anticipate slow degradations or unknown faults, especially in long-duration missions, and is not adequate for real-time applications due to its ground dependency. To overcome these shortcomings, recent works focus on embedding critical parts of the \gls{fdir} chain on-board, such as the monitoring of actuators of the \gls{aocs}, dramatically reducing reaction times \cite{zheng2024control,zhang2023anomaly}. %,guan2025high}. 
Yet, delegating anomaly detection and identification actions to a \gls{ml}-based onboard system raises a key question: how can we trust an algorithm that operates autonomously in orbit?

% For attitude-control actuators such as reaction wheels and CMGs, data-driven monitoring has progressed from thresholding to predictive and representation learning paradigms. These methods produce health indices and alarms that can be embedded within hierarchical FDIR, shortening reaction times and supporting graceful degradation.
% However, these models often behave as black boxes, providing limited insight into why a certain anomaly was detected or a recovery was triggered. For safety-critical applications like spacecraft control, such opacity is unacceptable: operators must be able to trace the reasoning behind autonomous decisions to validate their correctness and ensure accountability \cite{ribeiro2016should,lipton2018mythos}. Explainable Artificial Intelligence (XAI), therefore, becomes a foundational requirement—not an optional feature—for the deployment of trustworthy onboard autonomy \cite{nannini2024operationalizing,schumann2021explainable}. Explainable Anomaly Detection for spacecraft telemetry and model-level diagnostics are emerging to support verification, validation, and operator acceptance \cite{cuellar2024explainable-ad}.

Despite the success of black-box \glspl{dnn} for anomaly detection, their opacity is unacceptable given that they offer limited insights on why the anomalies are detected \cite{Ruff_2021ProcIEEE}. For safety-critical applications such as spacecraft control, operators must trace the reasoning behind autonomous decisions to validate their correctness and ensure accountability \cite{ribeiro2016should,lipton2018mythos}. Therefore, \gls{xai} becomes a foundational requirement, not an optional feature, for the deployment of trustworthy onboard autonomy \cite{nannini2024operationalizing,schumann2021explainable}, motivating the emergence of \gls{ead} for spacecraft telemetry and model-level diagnostics to support verification, validation, and operator acceptance \cite{cuellar2024explainable-ad}.

In this context, explainability, beyond post-hoc visualization, must be embedded into the decision loop. For on-board \gls{fdir}, this entails generating interpretable health indicators linked to physical quantities (e.g., current, vibration, or torque signatures) and conveying them to ground control together with confidence levels and causal evidence \cite{tang2022health,ciancarelli2022innovative,chen2021imbalanced}. A possible approach envisages the adoption of Physics-informed models, a promising path toward this goal. They combine data-driven adaptability with physically meaningful structure and constraints, improving robustness and interpretability \cite{wu2024physics,tang2022health}. Embedding such models in an autonomous on-board loop can facilitate human understanding of the system’s behavior, enable verifiable responses, and ultimately reinforce trust in satellite autonomy \cite{schumann2021explainable,nannini2024operationalizing}. Nevertheless, the deployment of these approaches on a platform with limited capabilities could represent a constraint to their usability.

Differently, this paper addresses these challenges and opportunities by introducing a framework that extracts high-level information directly from the internal activations of a fully data-driven model. Specifically, we consider the case of an advanced \gls{fdir} system that integrates a neural anomaly detector and enriches its decisions with semantic side information. The objective is to provide, for each alert, concise evidence describing the shape of the event and where the event is more evident. This auxiliary output facilitates the localization of the probable fault source. We also provide evidence that such semantic characterization can surface potential monitoring biases, revealing when and how the model’s internal focus may drift or over-emphasize specific channels, modes, or operating conditions.

As a case study, we implemented the proposed approach using a \gls{cnn}-based Autoencoder as an anomaly detector that processes telemetry data from four \gls{rw} units controlled by the \gls{aocs}. The network is coupled with the proposed non-neural processing chain applied to its intermediate activations to generate human-interpretable, compact descriptors. We emphasize that the framework is designed to transform raw detections into actionable and transparent evidence, supporting both on-board decision-making and post-event analysis.

The remainder of this paper is organized as follows. Section~\ref{sec: model} formalizes the proposed framework, detailing its three main stages: dimensionality reduction, statistical characterization, and semantic mapping. Section~\ref{sec: numerical evidence} applies the framework to an advanced FDIR use case within the AOCS, describing the reference scenario, the autoencoder-based anomaly detector, the synthetic anomaly models, and the adopted evaluation protocol. Section~\ref{sec:prel_res} summarizes the numerical evidence, while Section~\ref{sec: conclusion} discusses the main findings, current limitations, and outlines directions for future work focused on bias analysis.

\section{Mathematical model for peephole extraction}
\label{sec: model}

As a reference, we consider a general \gls{dnn} that accepts an input tensor $\img$ and produces an output $\ell(\img)$. This mapping is realized through a sequence of intermediate layers, each of which transforms the input activations $\actIn$ into the output tensor $\actOut$.

\begin{figure}
	\centering
	\includegraphics[width=0.5\textwidth]{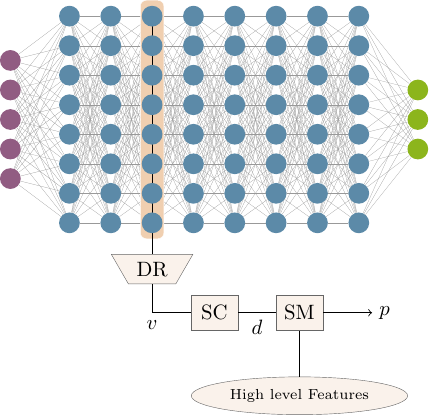}
	\caption{Block scheme of the proposed framework extracting \peephole vectors $\vec{p}$ from a target layer involving three stages: dimensionality reduction (DR), statistical characterization (SC), and semantic mapping (SM).}
	\label{fig:generic_nn}
\end{figure}

Our framework processes the intermediate activations of a \gls{dnn} to produce a \peephole vector, which represents a semantically annotated, low-dimensional encoding that enables high-level inspection of the network’s internal behavior. Starting from the model's activation in a single layer, our framework employs a three-stage non-neural processing pipeline as depicted in Fig.~\ref{fig:generic_nn}: \textit{i}) a dimensionality reduction  step that reduces the high-dimensional activation vector to a compact and equivalent representation of its information content, called \corevector $\coreVec$; \textit{ii}) a statistical characterization of those vectors to capture the structure of the \corevectors by clustering their typical positions in the $\coreVec$ signal space; \textit{iii}) a semantic mapping where the positions of the \corevectors with respect to the identified clusters are associated with a set of high-level, human-interpretable features, which facilitate the network’s output interpretation and are summarized in the final \peephole vector $\ph$.

 % The positions of the \corevectors relative to the identified clusters are then mapped to a set of high-level, human-interpretable features summarized in the final \peephole vector $\ph$.
As the main structure for the target layer, we focus on dense linear layers, which produce an output $\actOut$ according to an affine transformation of the input $\actIn$:
\begin{equation}\label{eq:preactivations}
	\actOut=\left[
		\weights | \biases 
	\right]
	\left[ \begin{matrix} 
		\actIn \\
		1
	\end{matrix}\right]=
    \affineTransform
    \left[ \begin{matrix} 
		\actIn \\
		1
	\end{matrix}\right],
\end{equation}
where $\weights$ and $\biases$ are a matrix and a vector containing the neurons' weights and the corresponding biases, while $\affineTransform$ remains implicitly defined. Most often, this kind of layer is followed by non-linear blocks, normalization, or aggregation layers\footnote{Note that \eqref{eq:preactivations} is a general form that models also other classes of layers, e.g., convolutional layers.}.

\subsection{Dimensionality reduction}\label{subsec: equivalent core activations}

To derive a reduced-dimensional representation of $\actOut$ that preserves the role of the neurons' parameters, we construct an auxiliary output $\coreVec$ based on the \gls{svd} \cite{Friedland2007} of the mapping in eq. \eqref{eq:preactivations}.
\begin{equation}\label{eq:SVD decomposition}
	\affineTransform=\SVDU\SVDS\SVDV^\top,
\end{equation}
where $\SVDU$ and $\SVDV$ are square orthonormal matrices containing left and right singular vectors, respectively, while $\SVDS$ is a rectangular diagonal matrix containing the singular values, conventionally sorted in non-increasing order.

Let $\SVDV'$ and $\SVDS'$ denote the matrices containing the first $\coreVecSize$ columns of $\SVDV$, and the first $\coreVecSize$ singular values of $\SVDS$, respectively. It is then possible to define a rank-$\coreVecSize$ approximation $\affineTransform' = \SVDU \SVDS' {\SVDV'}^\top$ such that the Frobenius norm $\|\affineTransform - \affineTransform'\|_F$ is minimized.

This property enables the definition of an auxiliary $k$-dimensional vector that preserves the effect of $\affineTransform$ on $\actIn$, defined as follows:
\begin{equation}\label{eq: corevector}
    \coreVec={\SVDV'}^\top
	\left[\begin{matrix}
		\act\\1
	\end{matrix}\right].
\end{equation}

\subsection{Statistical characterization}\label{subsec: low level feature extraction}
%\rev{actually, the input to the clustering stage is not just $\coreVec$, rather its standardized version wrt mean and stds of the training set}

This step aims to characterize, from a statistical perspective, how the vectors $\coreVec$ are distributed within their $k$-dimensional space. To this end, all vectors are normalized to have zero mean and unit variance, after which a clustering algorithm is applied to capture their distribution.

The statistical distribution of the normalized \corevectors is modeled using a \gls{gmm} \cite{dempster1977maximum}, which defines $\numClusters$ Gaussian components. Each component is associated with a membership probability function $\gamma_i: \realSet^\coreVecSize \rightarrow \realSet^+$, estimating the likelihood that a given $\coreVec$ belongs to the $i$-th cluster. The estimated likelihoods are then grouped in a  membership vector $\membership\in\realSet^{\numClusters}$ with components
\begin{equation}
\label{eq:d_clustering}
\membershipElement_i =
\frac{
\gamma_i(\coreVec)
}
{\sum_{j=0}^{C-1}\gamma_j(\coreVec)}
\end{equation}
For each cluster, \gls{gmm} gives as output a center $\means\in\realSet^\coreVecSize$, a covariance matrix $\covariance\in\realSet^{\coreVecSize\times \coreVecSize}$, and a weight $\clusterProb_i$ such that
\begin{equation}
\gamma_i(\coreVec)= \frac{\clusterProb_i}{\sqrt{(2\pi)^{\coreVecSize} \det(\covariance_i)}} \exp\left[-\frac{1}{2} (\coreVec - \means_i)^\top \covariance_i^{-1} (\coreVec - \means_i)\right]
\label{eq: membershipGMM}    
\end{equation}

Note that $\membership$ in \eqref{eq:d_clustering} is a $\numClusters$-tuple of non-negative real numbers that can be interpreted as probability assignments to $\numClusters$ mutually exclusive events.

%and in this case are the membership of the incumbent \corevector into each of the clusters.

%\rev{LC - I don't get $\membership\in\probSet^\coreVecSize$ shouldn't be $\membership\in\probSet^\numClusters$}

\subsection{Semantic mapping}
\label{subsec: high level feature extraction}
This final stage produces a vector that links the information passing through the target layer to a set of high-level, human-interpretable features.

This is achieved by associating the statistical characterization obtained from the clustering stage with a set of human-inspectable tags. Following the approach in \cite{liu2020explaining}, this relationship is modeled through two functions, $\corevecToCluster : \realSet^\coreVecSize \mapsto \{0, \dots, \numClusters - 1\}$ and $\corevecToTag : \realSet^\coreVecSize \mapsto \{0, \dots, \numTags - 1\}$, which map each \corevector respectively to a cluster and to a tag.

The connection between $\corevecToCluster(\coreVec)$ and $\corevecToTag(\coreVec)$ can be expressed in probabilistic form as
\begin{equation}
	\Pr\left\{\corevecToTag(\coreVec)=i\right\} = 
	\displaystyle\sum_{j=0}^{\numClusters-1}
	\Pr\left\{\corevecToTag(\coreVec)=i|\corevecToCluster(\coreVec)=j\right\}
	\Pr\left\{\corevecToCluster(\coreVec)=j\right\}    
\end{equation}
where $\membership$ can be interpreted as a probability distribution such that ${\rm Pr} \{\corevecToCluster(\coreVec) = j\} = \membershipElement_j$. The conditional probabilities $\empiricalPosterior_{i,j} = \Pr\left\{\corevecToTag(\coreVec) = i \mid \corevecToCluster(\coreVec) = j\right\}$ form a matrix that must be empirically estimated from data. Assuming that a set of examples is given such that for each instance in that set we know the conceptual label, we may compute the number $\empiricalPosteriorRowElement_{i,j}$ of joint events
\begin{equation}\label{eq:empiricalposterior_revised}
    \empiricalPosteriorRowElement_{i,j} = \left|\left\{\corevecToClass(\coreVec_t)=i \wedge \corevecToCluster(\coreVec_t)=j\right\}\right|
\end{equation}
and finally estimate the matrix $\empiricalPosterior$ as follows
\begin{equation}\label{eq:empiricalposterior}
    \empiricalPosteriorElement_{i,j}=\frac{\empiricalPosteriorRowElement_{i,j}}{\displaystyle\sum_{k=0}^{\numLabels-1}\empiricalPosteriorRowElement_{k,j}}
\end{equation}

Once estimated, $\empiricalPosterior$ is used to generate the final auxiliary output $\ph=\empiricalPosterior\membership$ that is named \peephole. In other words, the \peephole relates \gls{llf} learned by the \gls{dnn} with \gls{hlf}, which have meaning to humans through means of the empirical posterior knowledge encoded in $\empiricalPosterior$.

Note that two parameters control the \peephole extraction and tailor the proposed generic inspection mechanism to a specific application, optimizing its performance, namely the \corevectors dimension $\coreVecSize$, and number of clusters $\numClusters$ used to estimate the relationships between \glspl{llf} and \glspl{hlf} through $\empiricalPosterior$. 

\section{Numerical Evidence}\label{sec: numerical evidence}

\subsection{Reference scenario} \label{subsec: ref scen}

As discussed in Section \ref{sec: intro}, we focus on the detection of anomalies performed by an advanced \gls{fdir} controlling the operation of the \gls{aocs}. In particular, the reference case considered in this study is the detection of anomalous events in telemetry data from an on-board \gls{aocs} equipped with four \glspl{rw}, data collected during an ESA Earth Observation mission.

The \glspl{rw} provide continuous and precise attitude control by compensating for environmental disturbances. Each \gls{rw} is driven by an electric motor, whose sensor's telemetry is processed by the \gls{fdir} system. The focus of this work is on the first \gls{fdir} stage, detecting potential sources of non-ideal behavior.

Telemetry associated with the \gls{rw} includes parameters such as motor performance, rotational dynamics, and thermal conditions, offering a detailed characterization of their operational states. The dataset used in this study contains data from four \glspl{rw} (\gls{rw}0–\gls{rw}3), each providing four time series, for a total of 16 telemetry channels. These channels are processed by a detector designed to identify previously unseen anomalous events.

The dataset is segmented into chunks $\img \in \mathbb{R}^{16 \times 16}$, where the first dimension represents the window length, i.e., the number of samples per time series, and the second dimension corresponds to the number of time series included in each window. The subset of chunks representing the nominal system behavior is used for training and validation to determine the parameters of the detector. The training set consists of $8.5\times10^5$ samples, while validation and test sets include $2.1\times10^5$ and $1.2\times10^5$ samples, respectively.

For anomaly detection, we employ an Autoencoder, a neural architecture composed of two main networks. The first network, the encoder, compresses the input $\img$ into a latent representation $\vec{z}$, which resides in a lower-dimensional manifold capturing the most salient information. The second stage, the decoder, takes $\vec{z}$ as input and reconstructs an output tensor $\hat{\img}$ that aims to replicate the original input $\img$.

More specifically, the encoder stage consists of two convolutional blocks (layers with $38$ and $76$ filters, stride set to 1, and a kernel size of 3×3), followed by a flattening layer and a fully connected layer that produces the latent representation $\vec{z}\in\realSet^{256}$. The decoder mirrors the encoder architecture, employing transposed convolutional layers to reconstruct the input from the latent space. The model architecture comprises a total of approximately two million parameters.

The networks' parameters are trained by minimizing a loss function given by the \gls{mse} computed as the expectation of $\lVert \img - \hat{\img}\rVert_F^2$, where \gls{mse} acts also as an anomaly score. The reference model achieved a validation loss of $8.64\times10^{-5}$.

To simulate different fault conditions, normal instances are synthetically corrupted to generate anomalous samples \cite{WOMBAT}, denoted as $\img'$ in the following discussion. These anomalies are parameterized to maintain a \gls{snr} of $0~\rm{dB}$, ensuring a challenging detection scenario. Five distinct types of synthetic anomalies are introduced into the dataset. Each anomaly type is mathematically defined and controlled through parameters that regulate the severity of the perturbation. Specifically, given a single telemetry channel $\telemetry$ (column of $\img$):

\begin{description}
\item[Additive noise (GWN):]
Zero-mean white Gaussian noise is added to each element of the data instance $\telemetry$. The anomalous instance is defined as
$$
\telemetry' = \telemetry + a \noise,
$$
where $\noise \sim \mathcal{N}(\mat 0, \identity)$, with $\identity$ the identity matrix and $a \in \realSet_+$ determines the intensity of the injected anomaly.
\item[Offset:]
A constant offset with a random sign is uniformly applied to all elements of each data instance $\telemetry$. The anomaly is modeled as
$$
\telemetry' = \telemetry \pm a\mat 1,
$$
where $\mat 1$ is an all-ones vector.
\item[Impulse:]
A spike with a random sign is inserted at randomly selected positions in each telemetry. The position $j$ of the spike is uniformly sampled in $\{0, \dots, 15\}$.
$$
\telemetry' = \telemetry \pm a  \mat e_j,
$$
where $\mat e_j$ is the indicator (canonical basis) vector with a one at position $j$ and zeros elsewhere.
\item[Power Spectral Alteration (PSA):]
Nominal instances are corrupted by applying a random rotation matrix $\mat R_\theta$, i.e.,
$$
\telemetry' = \mat R_\theta \telemetry,
$$
where the rotation angle $\theta \in [0, \pi]$ controls the degree of alteration in the data and is set as $\theta = \arccos(1 - a^2 / 2)$.
\item \textbf{Step:}
As in the constant offset anomaly, the telemetry signal is perturbed by adding an offset with a random sign. In this case, however, the offset is applied only to a subset of the first $8$ or the last $8$ time samples within the current $\telemetry$.
$$
\telemetry' = \telemetry \pm a   \sum_{j=i}^{i+7}\mat e_j,
$$
with $i$ uniformly selected in $\{ 0, 8\}$.
\end{description}

More precisely, we generate two different versions of $\img'$. In a first case, we apply the same anomaly to all $16$ channels of $\img$, and we denote it as $\img'_I$. In a second scenario, we apply the same anomaly only to those channels associated with a specific \gls{rw}. The corresponding dataset is identified as $\img'_{II}$. In both cases, for each anomaly, the intensity $a$ is set such that the anomalies and normal signals have the same expected energy, i.e. $\expect[\| \telemetry' \|^2_2] = \expect[\| \telemetry \|^2_2]$, ensuring $\mathrm{SNR} = 0~\rm{dB}$.

The autoencoder, trained to minimize the \gls{mse} on nominal inputs, is subsequently used to generate anomaly scores that discriminate between nominal instances $\img$ and anomalous instances $\img'_I$ or $\img'_{II}$. To assess its performance, the distributions of the scores assigned to nominal and anomalous samples are compared using the $\mathrm{AUC}$ (Area Under the Receiver Operating Characteristic Curve) metric \cite{Fawcett_PATREC2006}. The $\mathrm{AUC}$ represents the probability that the score assigned to a nominal instance $\img$ is lower than that assigned to an anomalous instance $\img'_I$ or $\img'_{II}$. An $\mathrm{AUC}$ value approaching $1$ indicates a perfect detector, whereas $\mathrm{AUC}=0.5$ corresponds to a purely random predictor.

\begin{figure}
    \centering
    \includegraphics[width=0.8\linewidth]{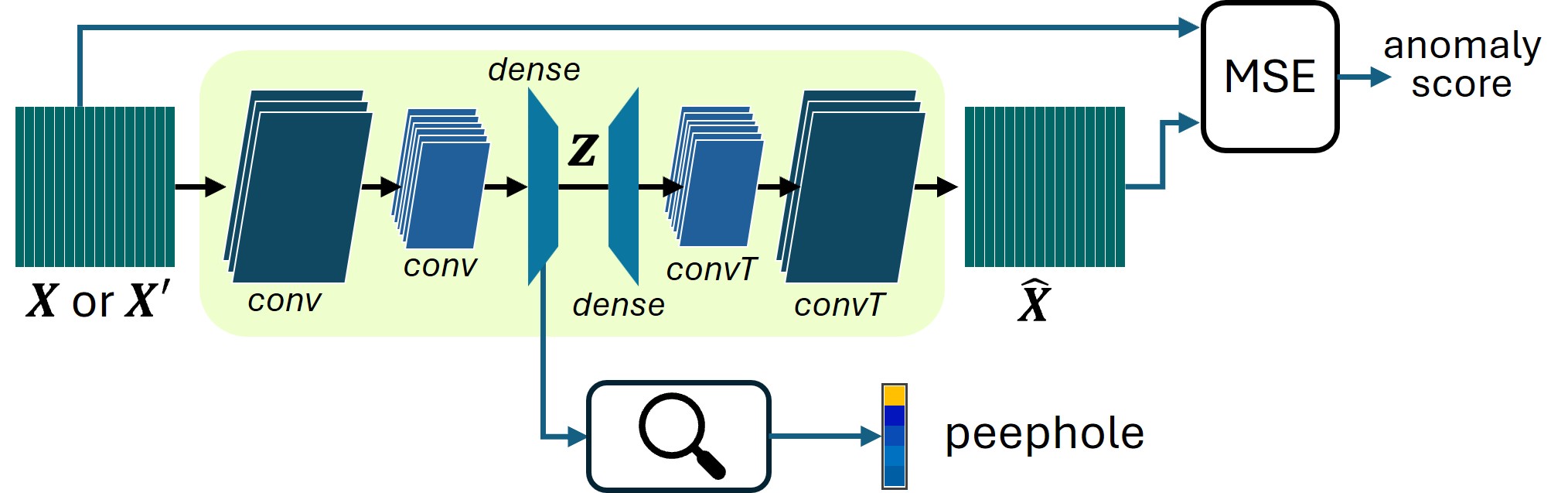}
    \caption{Block diagram of the autoencoder producing anomaly score, including semantic extraction from the encoder’s last layer}
\label{fig: AE}
\end{figure}

The obtained results, summarized in Tab.~\ref{tab:auc_anomalies}, indicate that the trained autoencoder performs as an almost perfect anomaly detector. Finally, a threshold is applied to the anomaly score to produce a binary output: a label of $0$ denotes predicted nominal behavior, while a label of $1$ indicates the detection of an anomalous or uncommon pattern. The threshold value is chosen such that the false positive rate remains below $0.001$. Each label of $1$ activates the block that, using the proposed framework, extracts a \peephole vector from the dense layer producing the latent representation $\vec{z}$ (see Fig.~\ref{fig: AE}).

\begin{table}[t]
\centering
\caption{AUC values for the five types of synthetic anomalies.}
\label{tab:auc_anomalies}
\begin{tabular}{@{}lccccc@{}}
%\begin{tabular}{lccccc}
\toprule
 & \,GWN\, & \,Offset\, & \,Impulse\, & \,PSA\, & \,Step\, \\
\midrule
$\img'_{I}$ & 1 & 1 & 1 & 1 & 1 \\
$\img'_{II}$ & 1 & 0.97 & 1 & 1 & 1 \\
\bottomrule
\end{tabular}
\end{table}

%\rev{This are the performances in terms of AUC (very high), we need a better name for the two scenarios and i think it would better to explain the two settings here the one where we apply it to all channels and the one where we apply it to a single RW}

\subsection{Semantic analysis}
\label{sec:prel_res}
The \peephole extractor described in Sec.~\ref{sec: model} is an interpretability tool that goes a step further in the direction of identification and isolation. Furthermore, it enhances a better understanding of how the autoencoder is performing the detection task.

In detail, we conduct some experiments that aim to verify the ability to extract information about both the type of anomaly and its location directly from the analysis of the internal activations of the layer that produces the latent-space representation, without introducing any additional neural classification blocks. This is done by using $\coreVecSize=50$ and $\numClusters=50$. The adoption of a neural classifier would lead to two undesirable effects: on one hand, it would reintroduce an opaque classification module requiring a dedicated training phase and offering no benefits in terms of explainability; on the other hand, it would risk making the detection and classification mechanism too computationally demanding to be deployed onboard.

To inspect the capability of the proposed framework to extract semantic information from the internal activation, avoiding the need for a further neural block, we analyze the \peepholes generated respectively from $\img'_I$ and $\img'_{II}$ under the condition that such instances are marked as anomalous by the autoencoder-based detector. 

Regarding the former case, starting from the anomalies described in Section~\ref{subsec: ref scen}, we generate corrupted versions of both the validation and test sets, each containing the same number of instances for every anomaly type. These anomaly types also serve as the \glspl{hlf} used to derive the matrix $\empiricalPosterior$ in~\eqref{eq:empiricalposterior}.

The second scenario considers $\img'_{II}$ as the anomalous instances. In this case, anomalies are injected only into the channels associated with a specific \gls{rw} target, which defines the second class of \glspl{hlf}. As in the previous scenario, an analysis of the anomaly types can still be carried out. Moreover, it is possible to generate distinct sets of \peepholes based on the particular \gls{rw} where the anomalies have been injected, treating the \gls{rw} as a \gls{hlf}. Note that this enables identifying which part of the monitored subsystem is causing the anomaly.

\begin{figure}[tp]
    	\begin{subfigure}[b]{0.47\linewidth}
    		\centering
            \includegraphics[width=0.8\linewidth]{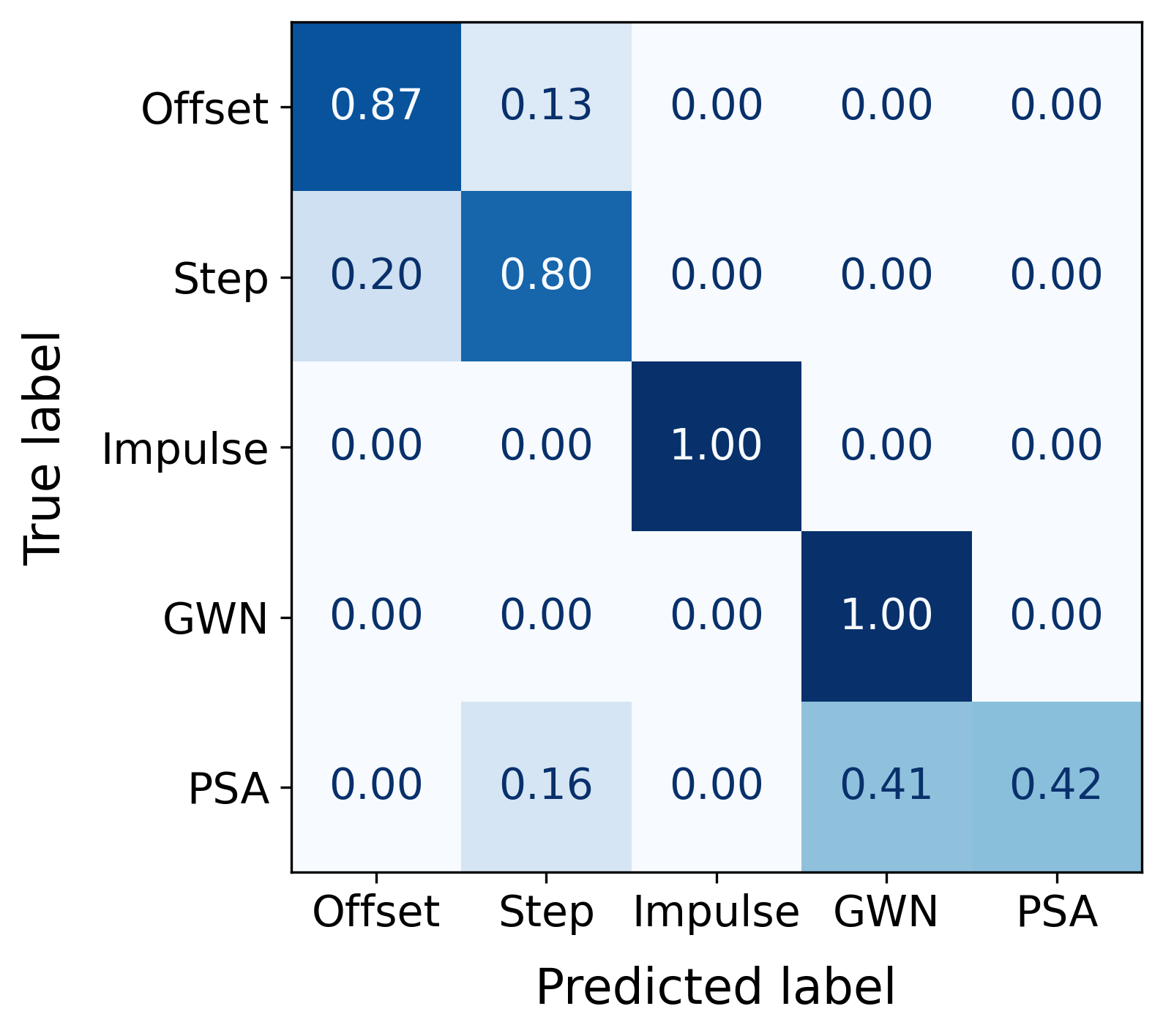}
            \caption{Perturbation applied to all channels.}
            \label{fig: anomaly id all}
    	\end{subfigure}
        \begin{subfigure}[b]{0.47\linewidth}
    		\centering
            \includegraphics[width=0.8\linewidth]{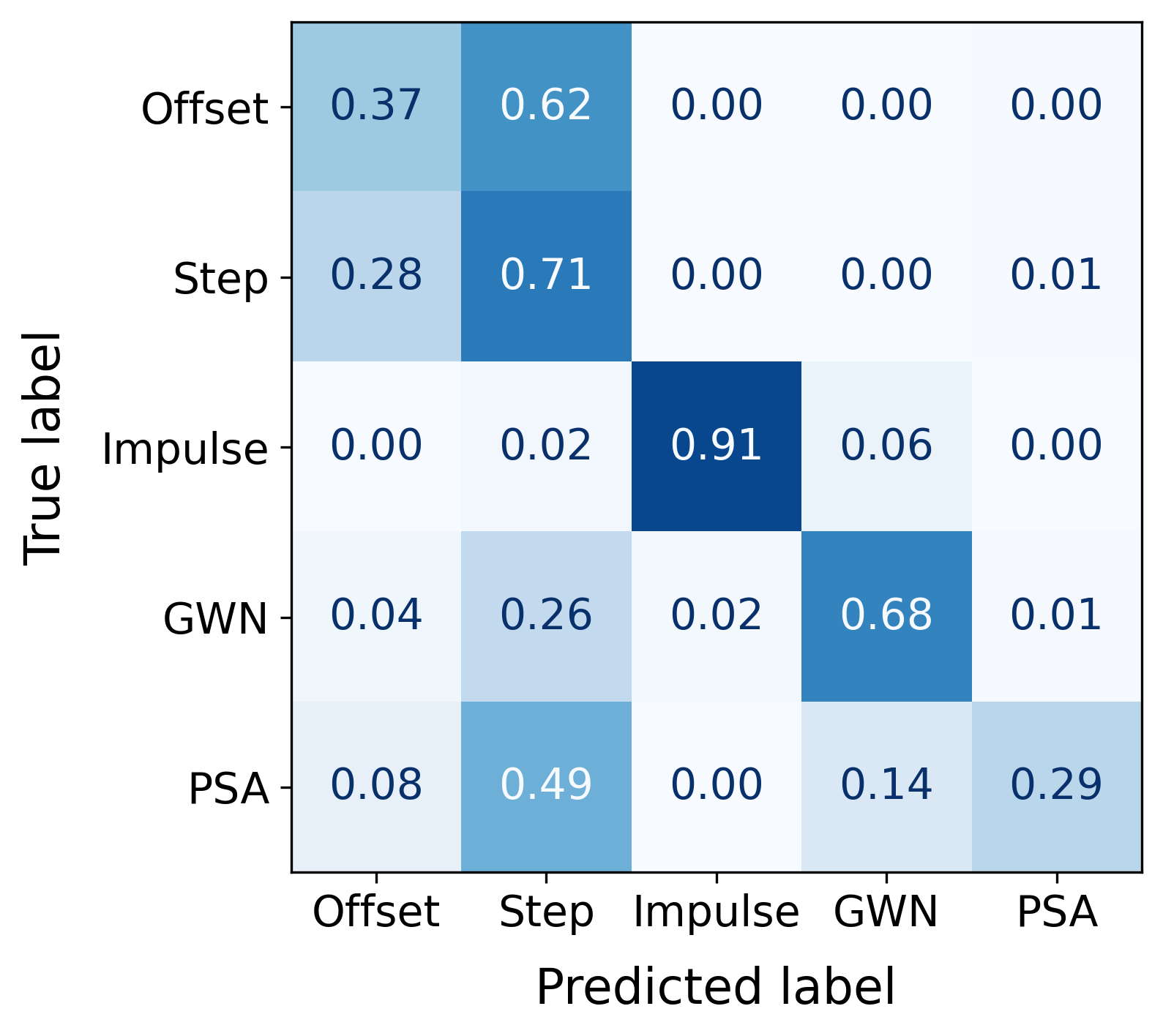}
            \caption{Perturbation applied to a single \gls{rw}.}
            \label{fig: anomaly id rw}
    	\end{subfigure}
	\caption{Anomaly identification when applied to all channels or to a single \gls{rw}.}
	\label{fig: anomaly id}
\end{figure}

Concerning the ability to identify the specific anomaly type triggered by the autoencoder, the corresponding confusion matrices for the two considered scenarios are presented in Fig.~\ref{fig: anomaly id}. Each matrix entry represents the estimated probability of classifying an anomaly of the type indexed by the row as the one indexed by the column, with darker shades denoting higher probabilities. Reported results evidence that the \peephole is well aligned with this first task, the autoencoder tends to separate anomalies with different shapes except for the PSA case, where the detection partially overlaps with GWN and Step. In the second scenario, when anomalies corrupt the telemetries associated with one \gls{rw}, although the ability to distinguish the anomaly shapes is less pronounced than in the first scenario, a tendency to group semantically similar anomalies can be observed. In particular, GWN and PSA appear partially overlapped, as do the constant and step anomalies.

\begin{figure}[h!]
    \centering
    % First row
    \begin{subfigure}[b]{0.32\linewidth}
        \centering
        \includegraphics[width=\linewidth]{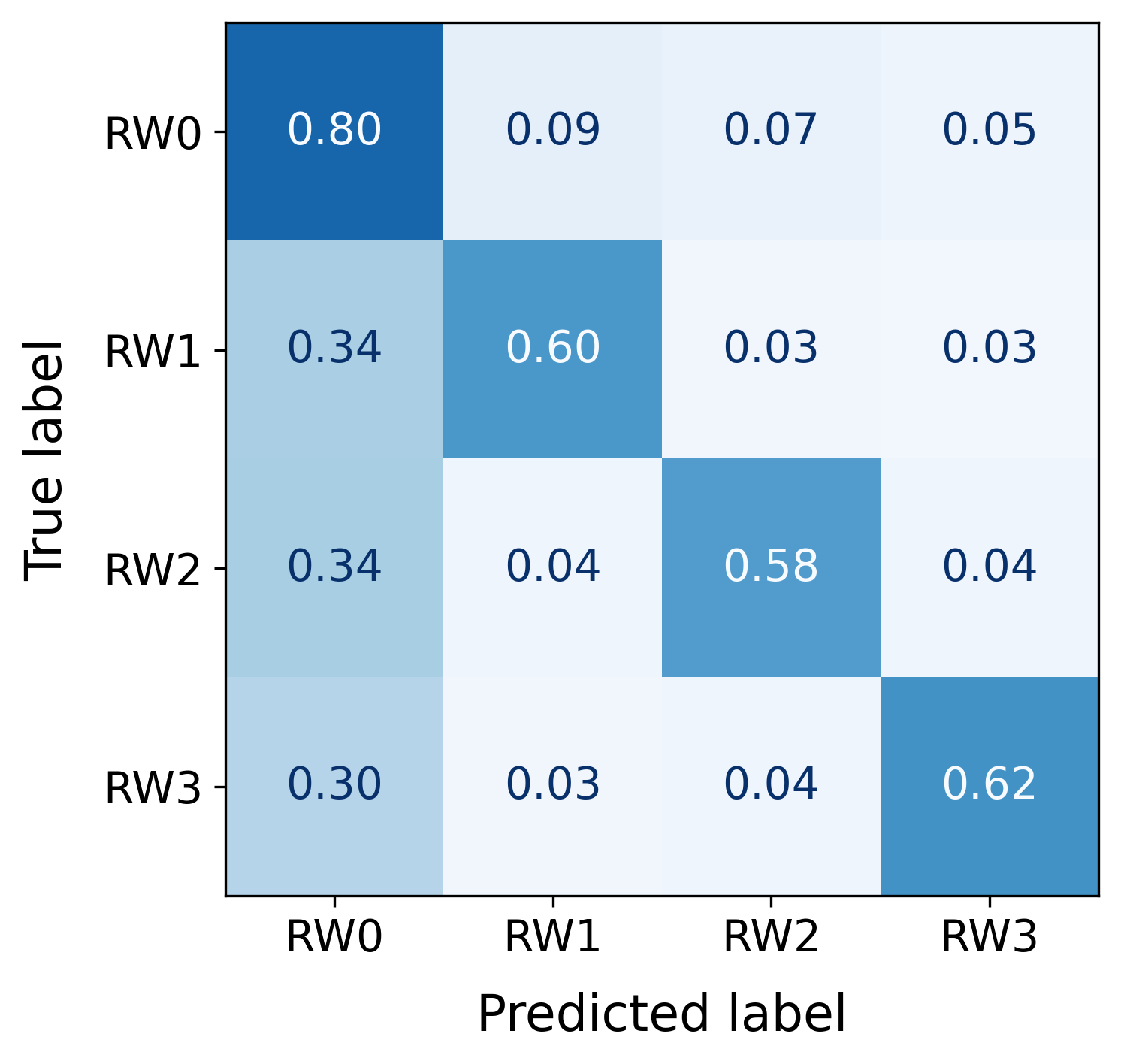}
        \caption{Overall}
        \label{fig:RW_overall}
    \end{subfigure}
    \hfill
    \begin{subfigure}[b]{0.32\linewidth}
        \centering
        \includegraphics[width=\linewidth]{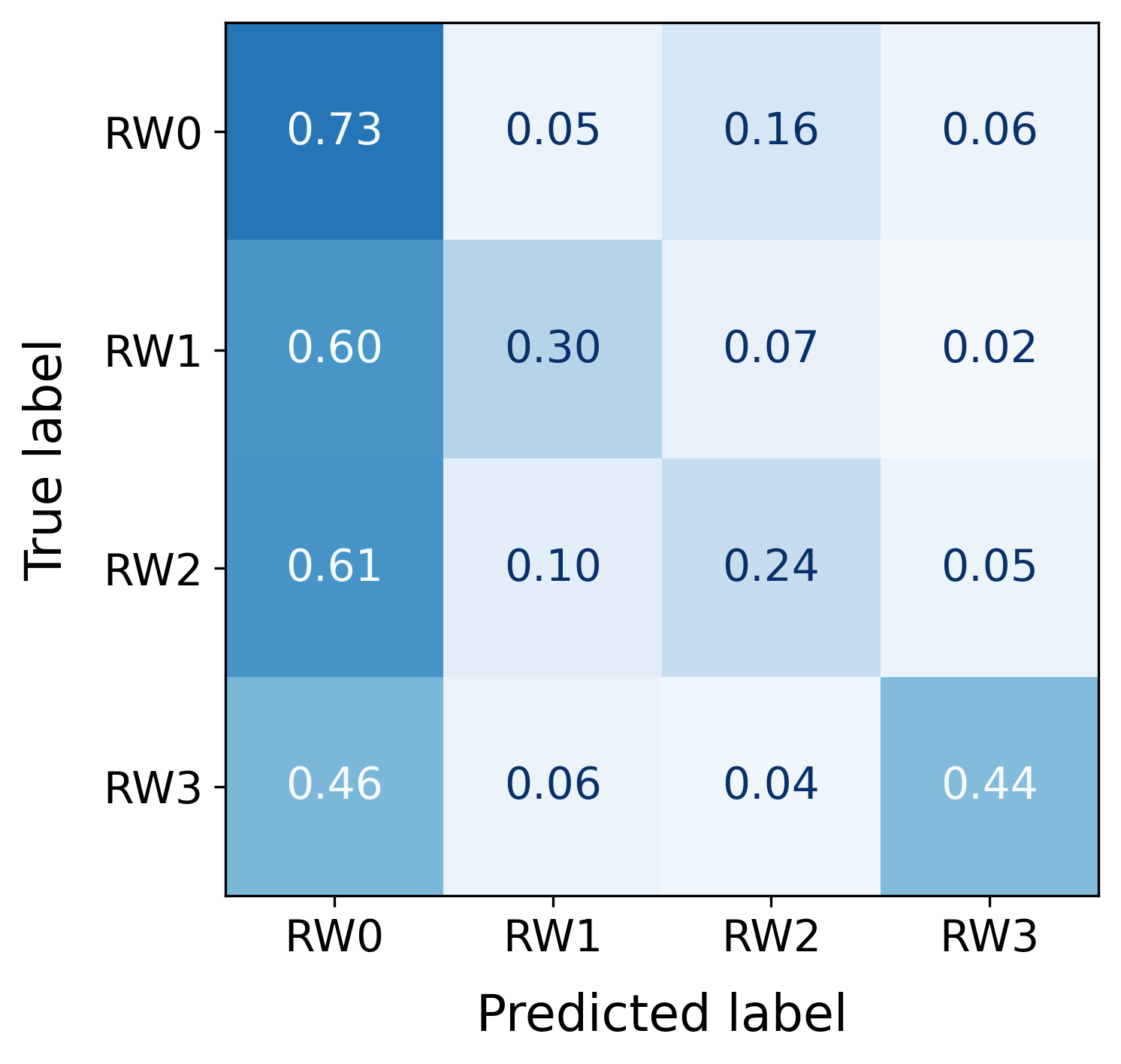}
        \caption{Offset}
        \label{fig:RW_offset}
    \end{subfigure}
    \hfill
    \begin{subfigure}[b]{0.32\linewidth}
        \centering
        \includegraphics[width=\linewidth]{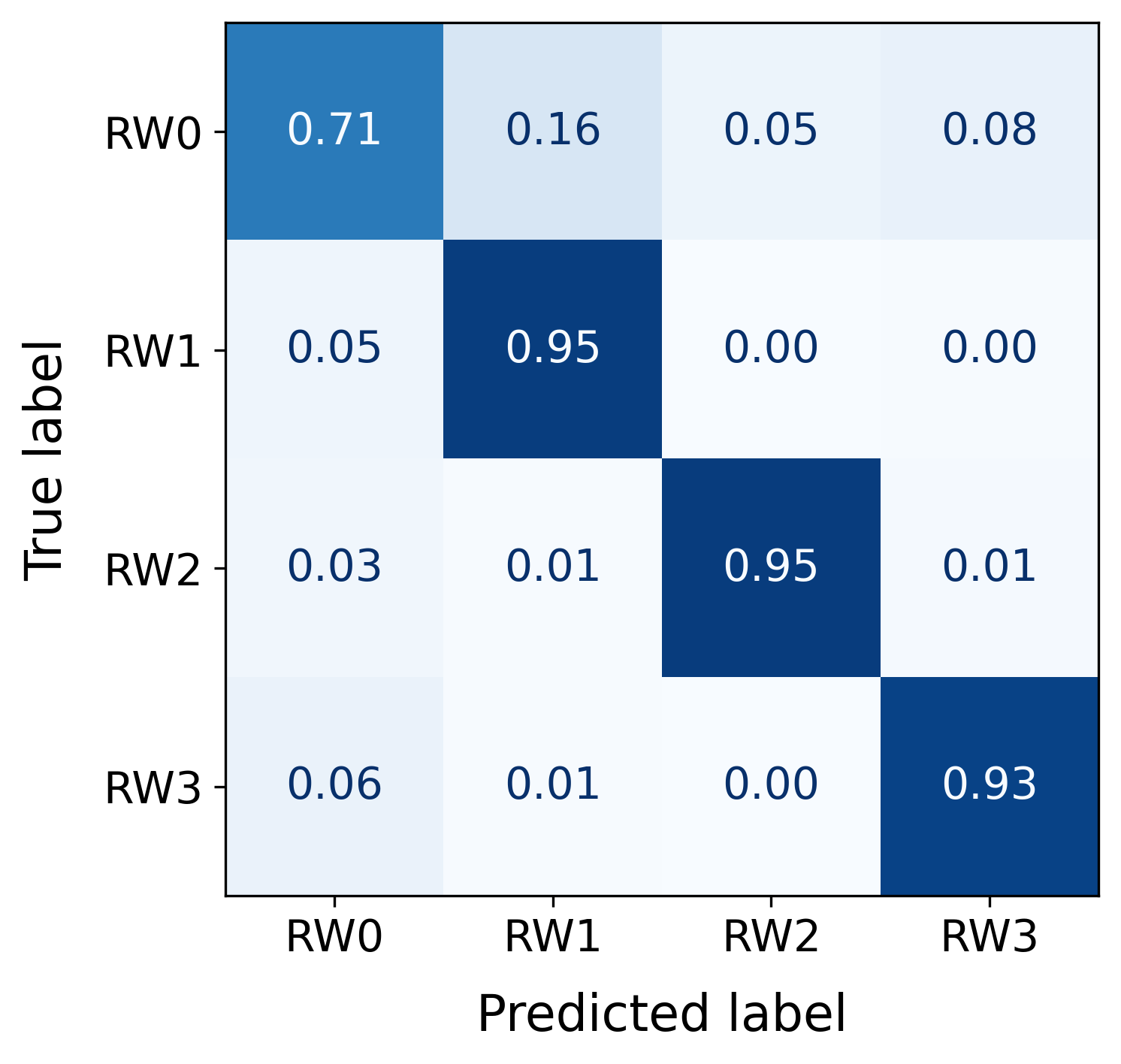}
        \caption{GWN}
        \label{fig:RW_GWN}
    \end{subfigure}

    \vspace{0.4cm}

    % Second row
    \begin{subfigure}[b]{0.32\linewidth}
        \centering
        \includegraphics[width=\linewidth]{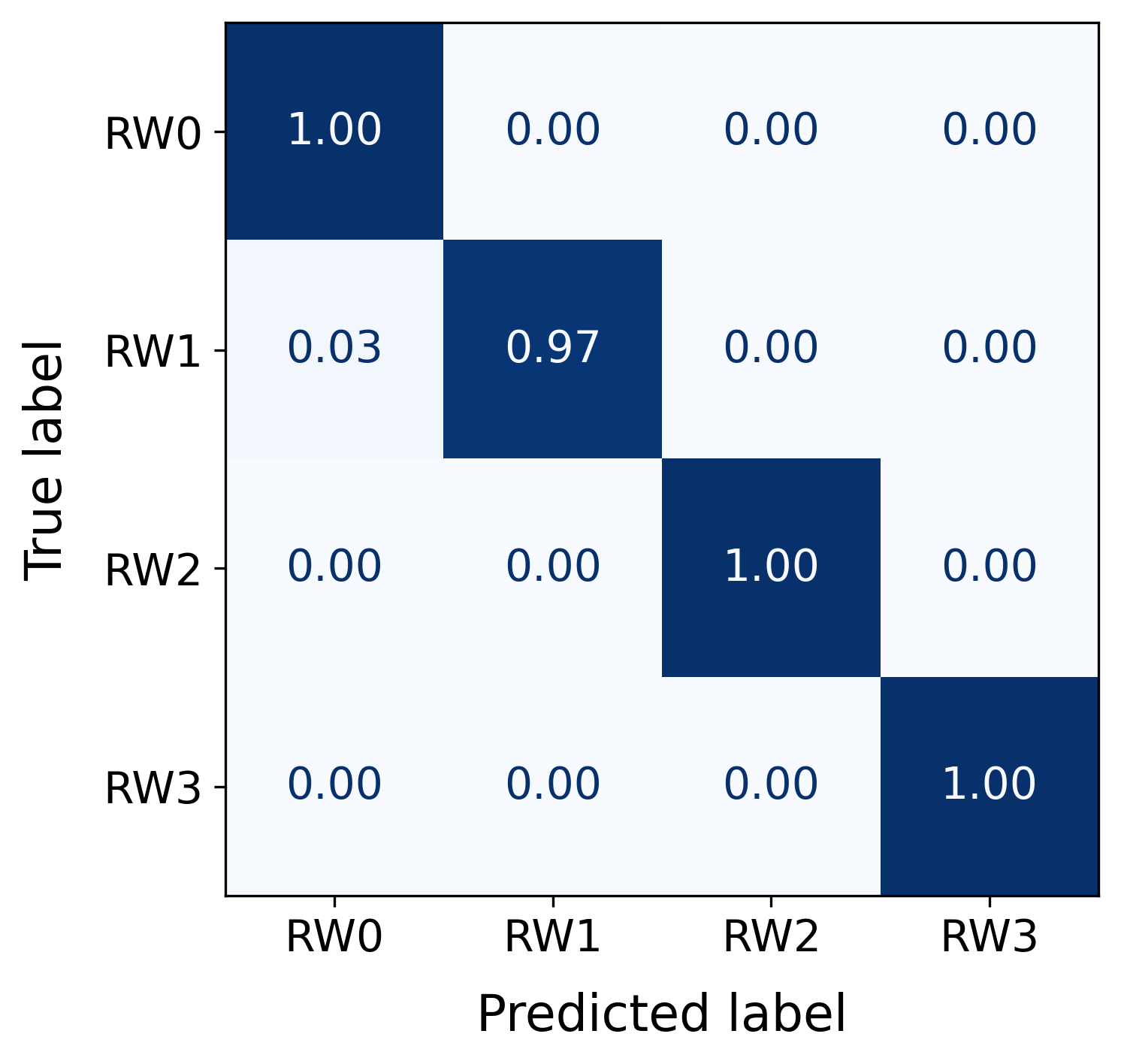}
        \caption{Impulse}
        \label{fig:RW_impulse}
    \end{subfigure}
    \hfill
    \begin{subfigure}[b]{0.32\linewidth}
        \centering
        \includegraphics[width=\linewidth]{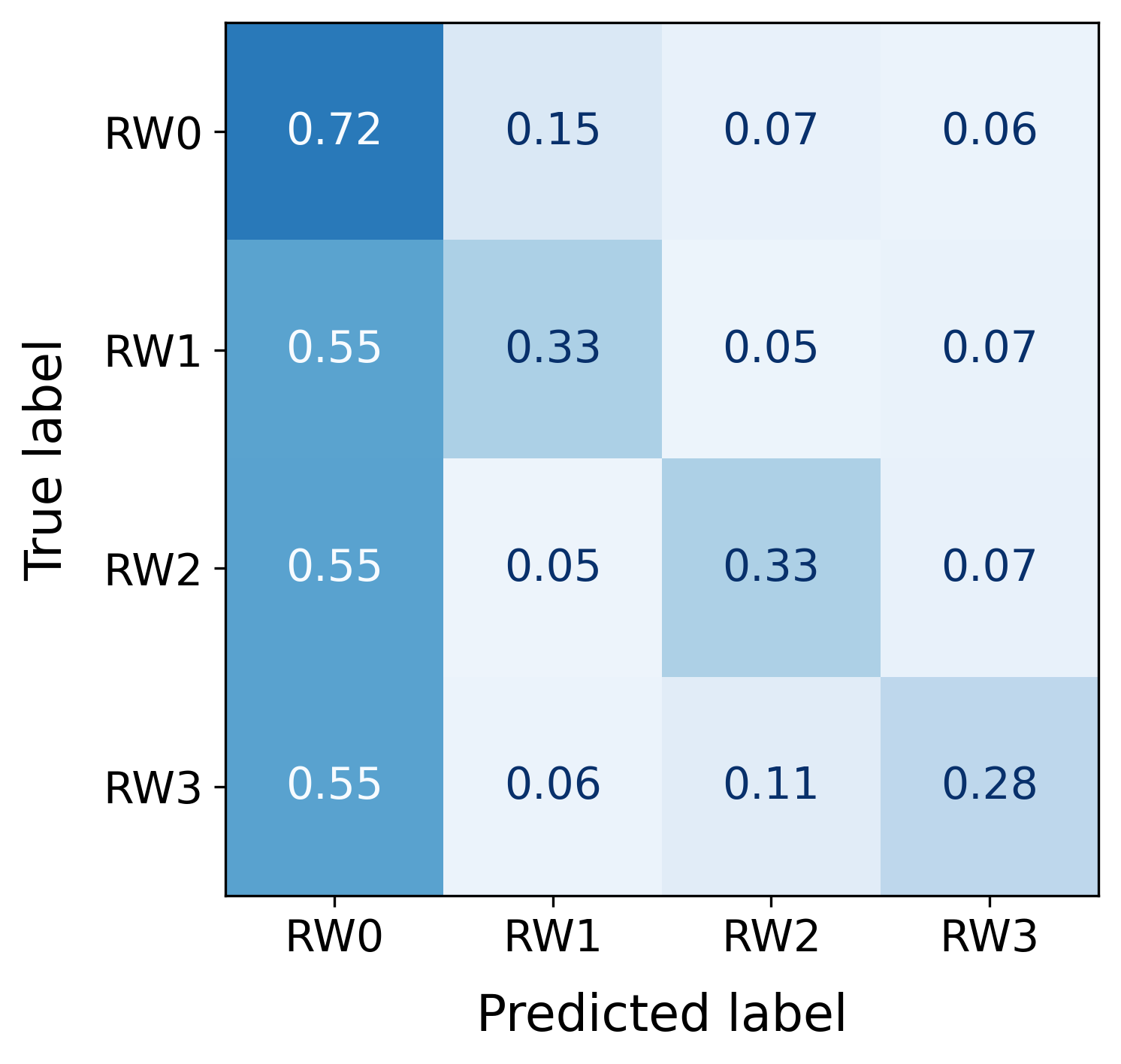}
        \caption{PSA}
        \label{fig:RW_PSA}
    \end{subfigure}
    \hfill
    \begin{subfigure}[b]{0.32\linewidth}
        \centering
        \includegraphics[width=\linewidth]{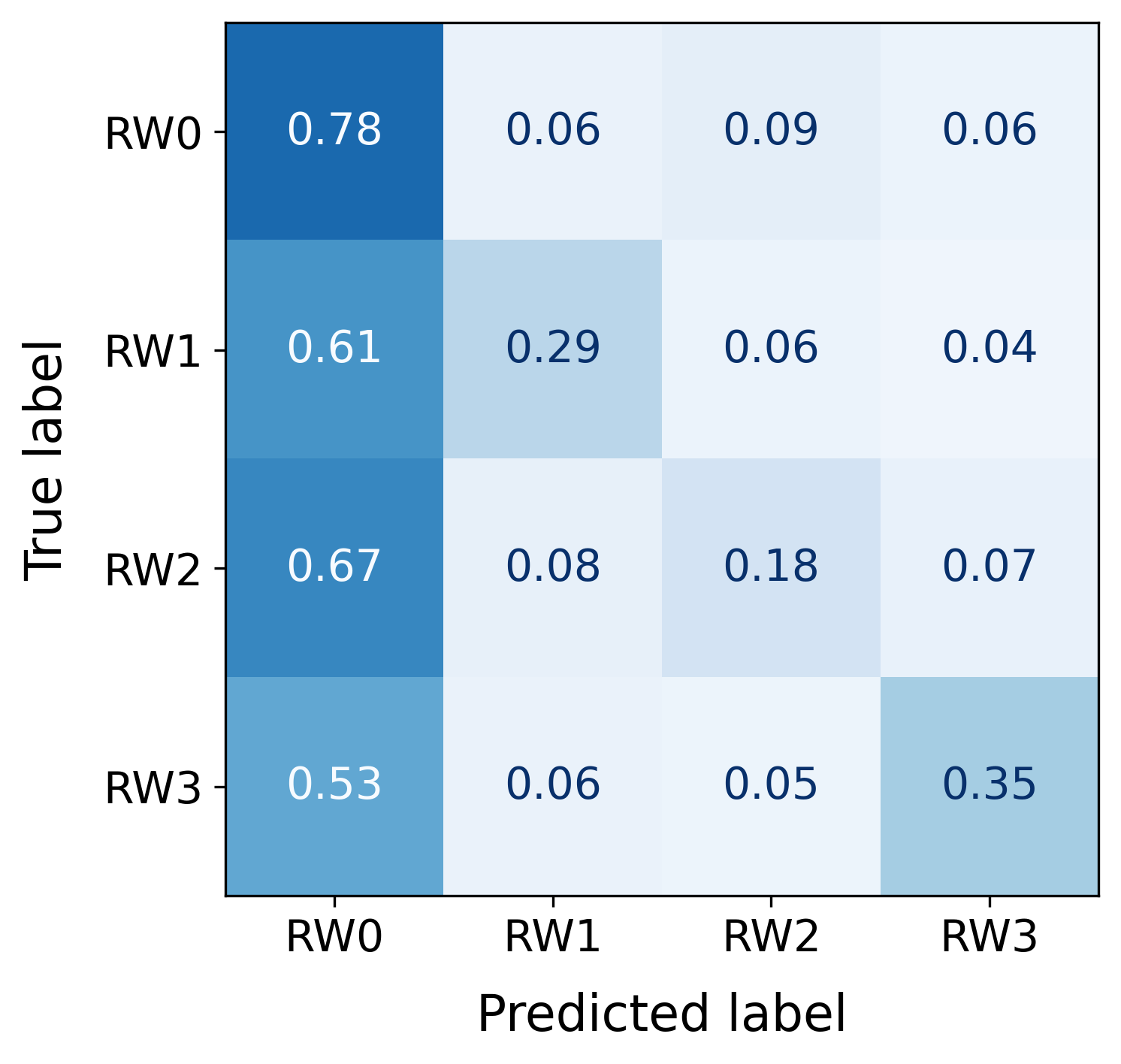}
        \caption{Step}
        \label{fig:RW_step}
    \end{subfigure}

    \caption{Confusion matrices for RW identification using Peephole for five anomalies.}
    \label{fig:RW_id}
\end{figure}

By considering the identification of the perturbed \gls{rw} through \peepholes in the second scenario, the results are reported in Figure \ref{fig:RW_id}(a). The obtained results show that there is a slight differentiation among the different \gls{rw}; however, a strong bias towards the first \gls{rw} is visible. This result highlights a bias exhibited by the trained autoencoder. This preliminary result was further investigated by analyzing whether the identification of the \gls{rw} was influenced by the typology of applied anomaly. As the confusion matrices presented in Fig.~\ref{fig:RW_id}(b)--(f) show, the bias towards \gls{rw}0 is not uniformly visible on all anomalies. In particular, impulse and GWN perturbation are easily identified on all $4$ \gls{rw}. The bias towards \gls{rw}0 is instead particularly evident on the remaining families of anomalies.


\begin{figure}
    \centering
    \includegraphics[width=\textwidth,height=\textheight,keepaspectratio]{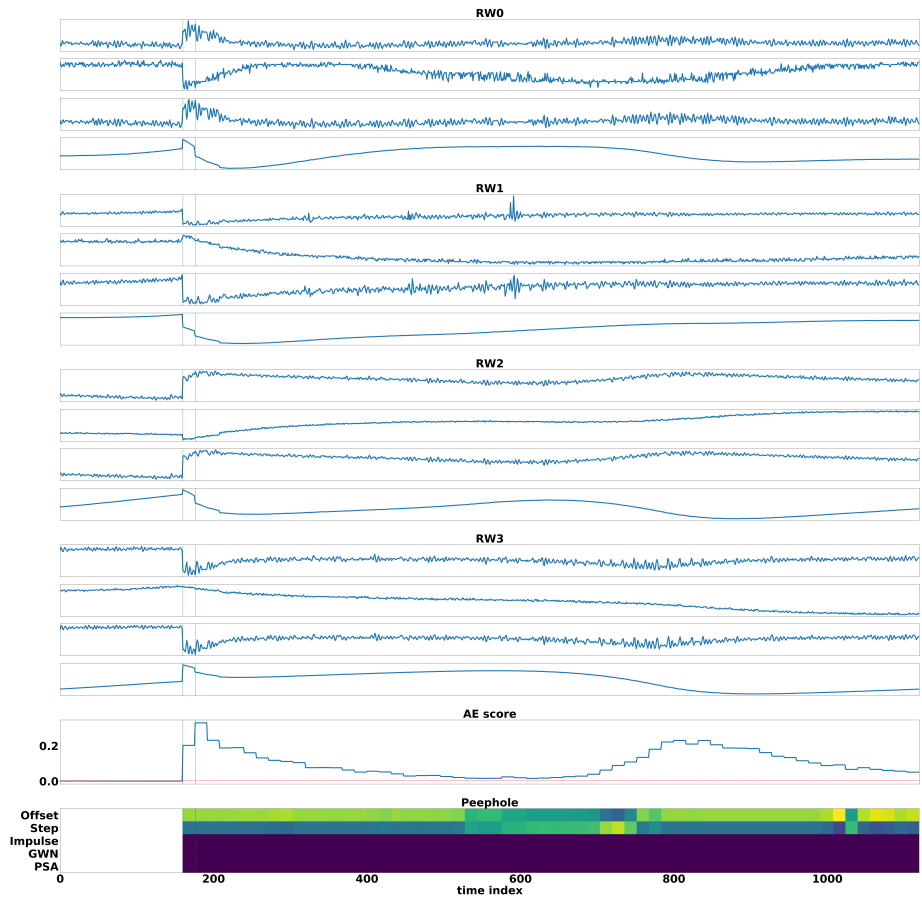}
    \caption{Telemetries of a true anomalous event along with the autoencoder's score and the corresponding \peepholes vectors visualized as a heatmap.}
\label{fig: TA}
\end{figure}

In Figure \ref{fig: TA} we report the application of the proposed framework for the identification of a real anomaly present in the test set and identified by domain experts. In this last case, the sequence of \peephole generated in correspondence with a real anomaly found in the dataset is presented. Specifically, Fig.~\ref{fig: TA} first shows the 16 telemetry signals, followed by the Autoencoder’s score profile, and finally a visual representation of all the computed \peephole vectors. The latter highlights a semantic signature of the anomaly, which is confirmed by the telemetry data, displaying a trend consistent with an offset- or step-type anomaly.

\section{Conclusion}\label{sec: conclusion}

This work introduces a framework for explainable onboard anomaly detection in autonomous spacecraft, based on low-dimensional, semantically annotated encoding vectors derived from neural activations, named \peephole. Applied to a convolutional autoencoder trained on real reaction wheel telemetry, the approach enables interpretable insights into the detected anomalies.

The \peephole allows the identification of the type and origin of anomalies directly from the latent representation, without additional neural classifiers, thus maintaining transparency and low computational cost. Preliminary results also revealed its potential for bias inspection, highlighting how detection behavior may vary across reaction wheels and anomaly types.

\section{Acknowledgment}
This study was partially carried out within the FAIR - Future Artificial Intelligence Research and received funding from the European Union Next-Generation EU
% (PIANO NAZIONALE DI RIPRESA E RESILIENZA (PNRR) – MISSIONE 4 COMPONENTE 2, INVESTIMENTO 1.3 – D.D. 1555 11/10/2022, PE00000013).
(Piano Nazionale di Ripresa e Resilienza (PNRR) – Missione 4 Componente 2, Investimento 1.3 – D.D. 1555 11/10/2022, PE00000013, and PE7 - CUP J33C22002810001, D.D. 341 15/03/2022 PE00000014).
This manuscript reflects only the authors’ views and opinions, neither the European Union nor the European Commission can be considered responsible for them.
\bibliographystyle{unsrt}
\bibliography{biblio}

\end{document}